# Binary Tree based Chinese Word Segmentation[1]

ZHANG Kaixu, WANG Can, SUN Maosong

State Key Lab of Intelligent Technology and Systems

Tsinghua National Laboratory for Information Science and Technology

Department of Computer Science and Technology

Tsinghua University, Beijing, 100084, China P.R.

**Abstract**: Chinese word segmentation is a fundamental task for Chinese language processing. The granularity mismatch problem is the main cause of the errors. This paper showed that the binary tree representation can store outputs with different granularity. A binary tree based framework is also designed to overcome the granularity mismatch problem. There are two steps in this framework, namely tree building and tree pruning. The tree pruning step is specially designed to focus on the granularity problem. Previous work for Chinese word segmentation such as the sequence tagging can be easily employed in this framework. This framework can also provide quantitative error analysis methods. The experiments showed that after using a more sophisticated tree pruning function for a state-of-the-art conditional random field based baseline, the error reduction can be up to 20%.

## Introduction

Each Chinese word consists of one or more characters. But there are no delimiters between characters in the sentences to indicate words. Since words are the basic units for many natural language processing tasks, Chinese word segmentation (CWS) is considered as a fundamental task for Chinese language processing. Languages such as Japanese, Thai and Vietnamese have similar problems.

The state-of-the-art methods treat the CWS as a character sequence tagging task like the POS-tagging task. A tag indicates the position of the corresponding character in the word.

We point out that the state-of-the-art methods suffer from a problem called the **granularity mismatch**. In CWS, it means that the granularity of the output is hard to perfectly match the granularity of the gold standard (the correct result). Without such problem, the performance is claimed to be increased by Li and Sun[5].

For example, the string 老树(*old tree*) in the MSR corpus in SIGHAN bake-off 2005[1] is considered as two words, 老(*old*) and 树(*tree*), whereas the string 老人(*old people*) in the same corpus is a single multi-character word consists of two morphemes, 老(*old*) and 人(*people*). Similar examples are quite common in any Chinese corpus, and cause most of the errors for CWS models.

[1] Received: 2011-04-##

Supported by the National Natural Science Foundation of China (No. ####).

*To whom correspondence should be addressed. Tel: ####; E-mail: sunmaosong@gmail.com

The explanation of this language phenomenon is that the boundary between the Chinese morphology and the syntax is not clear. The Chinese morphology and syntax share rules and even units (many morphemes such as "老" can also be used as free words). Sometimes it is hard to determine that whether a structure is morphological or syntactical (However, the CWS model has to do this). Historically, a large number of multi-character words in modern Chinese are used to be phrases in ancient Chinese.

In order to represent the structures of the unclear part between the morphology and syntax, and to overcome the granularity mismatch problem, we propose the binary tree representation, and a binary tree based CWS framework. In this framework, the CWS is divided into two steps, namely tree building and tree pruning. Fig. 2 shows the data flow in this framework.

In Step 1, the raw Chinese sentence is parsed to a binary tree (see Fig. 4 for an example) based on a simple tree building function. After this step ``老树'' and ``老人'' will have similar binary trees:

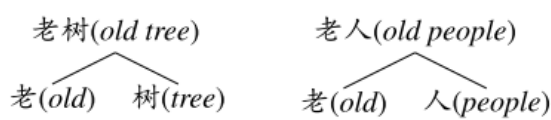

Fig. 1

In the tree building step, it is no need to determine that whether the structures are morphological or syntactical. Step 2 is designed to focus on the granularity problem. In Step 2, the tree is pruned based on a tree pruning function. The leaves of the pruned tree (see Fig. 6 for an example) form the output.

This binary tree based framework is with several benefits. First, it is a simple framework that can employ many previous CWS methods such as the dictionary based methods, the association measure based methods, and the sequence tagging based methods. Though in the framework we build trees for sentences, the training data is not needed to have any extra manually annotations. We will describe the two steps of the framework in Section 2.

Second, it provides quantitative error analysis methods which are described in Section 3. From such analysis, we see that the granularity mismatch problem is the primary cause of error for both "mono corpus" CWS and cross-corpus CWS[2].

The tree pruning step in our framework provides us a way to focus on the granularity problem. More sophisticated method can be employed in this step. We illustrated this idea in Section 4 by using an SVM-based model for the tree pruning. The experiments in Section 5 show that the errors reduced up to 20% comparing to the state-of-the-art CRF-based baseline.

---

[2] The definitions of Chinese word are not consistent between different corpora. The performance will drop a lot if we do cross-corpus CWS (i.e., train the CWS model from one corpus but test it on another one). This is also a research issue for CWS.

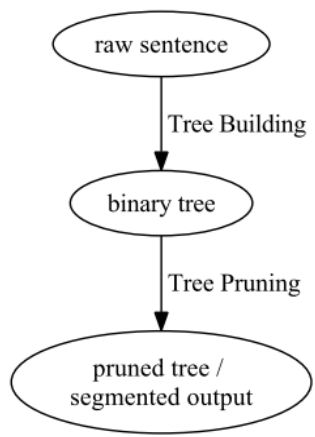

**Fig. 2　The data flow of the binary tree based CWS framework**

# 1 Related Work

Dictionary Matching algorithms such as the forward maximum matching algorithm and the backward maximum matching algorithm are greedy algorithms. The forward maximum matching algorithm finds the longest word in the dictionary that the input sentence starts with, and do this matching recursively for the rest of the sentence. The backward maximum matching algorithm does similar process just from the end of a sentence.

These algorithms will fail if the sentence contains any out-of-vocabulary (OOV) words (Words that do not appear in the training data or in the used dictionary are called OOV words. Otherwise they are called IV words).

Association measures such as the pairwise mutual information (PMI) and the $t$-test are used for CWS \cite{sun_chinese_1998}. These methods treat Chinese words as the "character collocations" and use collocation extraction methods to find them.

Xue \shortcite{xue:2003_j} proposed a character sequence tagging framework which is like the POS-tagging task. In such framework, the input is a raw Chinese sentence $\boldsymbol{s}$, which can be seen as a sequence of characters $c_i$.

$$\boldsymbol{s} = \overline{c_1 \cdots c_n}$$

The output of the character sequence tagging is a sequence $\boldsymbol{o}$ of labels $t_i$ corresponding to the input characters.

$$\boldsymbol{o} = \overline{t_1 \cdots t_n}$$

where $t_i \in \{\text{B, M, E, S}\}$. The tag B / M / E indicates the corresponding character is at the beginning / middle / end of a multi-character word. The tag S indicates the corresponding character is a single character word.

For example, if the gold standard result for the input ``材料利用率高" is "材料 利用率 高", the corresponding correct tag sequence will be B E B M E S. And in MSR corpus "老树" is tagged as S S, and ``老人" is tagged as $\textsf{B~E}$.

This character sequence tagging framework can be implemented by a CRF\cite{peng_chinese_2004} model, a perceptron\cite{gao_chinese_2005} or other models. Some reranking methods \cite{jiang_word_2008} are proposed to adjust the output.

The sequence tagging methods are considered to have the ability to identify the OOV words and make use of existent dictionary \cite{kruengkrai_error-driven_2009}. We will show that they still suffer from the granularity mismatch problem.

Some tree based methods for Chinese word segmentation were proposed\cite{zhao_character-level_2009,liu_information_2008} in order to represent the morphological or syntactical structure for better CWS. But these methods are hard to be applied for they need the training data to be extra manually annotated.

## 2 Method

### 2.1 Tree Building

The first step of our framework is to build a binary tree for an input sentence $\overline{c_1 \cdots c_n}$. The process can be simply based on a single function $b(c_i)$, which gives the confidence that there is a word boundary between $c_i$ and $c_{i+1}$.

The function $b(c)$ could be derived from various previous works.

For a PMI-based method\cite{sun_chinese_1998} which is an association measure based method, we can define the function $b_{\text{PMI}}(c)$ as:

$$b_{\text{PMI}}(c_i) = \log \frac{P(c_i c_{i+1})}{P(c_i)P(c_{i+1})}$$

For the CRF-based method, we define this function $b_{\text{CRF}}(c)$ as the marginal probability that:

$$b_{\text{CRF}}(c_i) = P(t_i = \text{S} \vee t_i = \text{E}|\mathbf{s})$$

The algorithm to building the binary tree based on this function is described as the function [x] as the pseudo code.

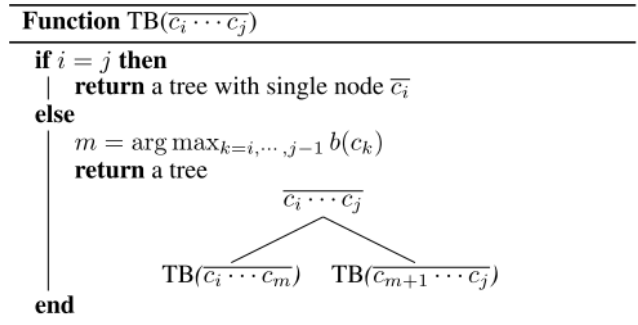

Fig. 3

This function \ref{function:tb} recursively splits the sequence $\overline{c_i \cdots c_j}$ at $c_m$ which has a maximum $b(c)$, until all the leaves are single characters.

An example of the binary tree is showed in Fig. \ref{fig:tree}. Notice that this is a tree for the entire sentence, but it is not a tree used to represent the morphological or syntactical structure. It represents the structures of the unclear part between the Chinese morphology and syntax.

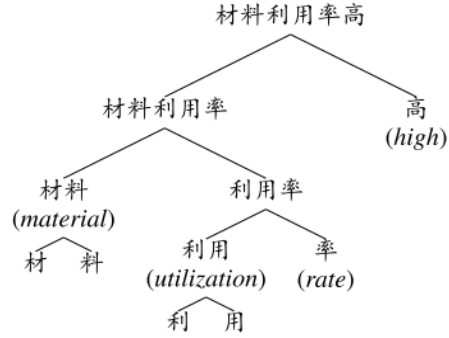

Fig. 4　**An binary tree for the phrase 材料利用率高~({\it the material utilization rate is high}) in the SIGHAN bake-off 2005 MSR corpus**

## 2.2 Tree Pruning

The second step is to prune the binary tree, which focus on the granularity. The leaves of the pruned tree form the segmentation output. The example of a pruned binary tree is showed in Fig. \ref{fig:pruned-tree}.

The pruning can also be applied by a single function $p(\overline{c_i \ldots c_m}, \overline{c_{m+1} \ldots c_j})$, where $\overline{c_i \ldots c_m}$ and $\overline{c_{m+1} \ldots c_j}$ are the roots of two subtrees of the node $\overline{c_i \ldots c_j}$. The function returns 1 if the subtrees of the node $\overline{c_i \ldots c_j}$ should be pruned, or 0 if not.

This binary function can be based on a threshold $t$ and the same $b(c)$ function used for the tree building:

$$p_{\text{threshold}_t}(\overline{c_i \ldots c_m}, \overline{c_{m+1} \ldots c_j}) = \begin{cases} 1, & b(c_m) < t; \\ 0, & \text{otherwise.} \end{cases}$$

([x])

This is a trivial pruning function. When we set $t$ to 0.5 and set the $b(c)$ to $b_{\text{CRF}}(c_i)$, the output is the same as the output by the original CRF-based method.

Word dictionary can be easily employed in the pruning step as another pruning function:

$$p_{\text{dictionary}}(\overline{c_i \ldots c_m}, \overline{c_{m+1} \ldots c_j}) = \begin{cases} 1, & \overline{c_i \ldots c_j} \text{ is in the dictionary;} \\ 0, & \text{otherwise.} \end{cases}$$

([x])

Notice that the values of particular pruning function may contain conflicts. For example, for the binary tree in Fig. \ref{fig:tree}, we may have $p(材料, 利用率) = 1$ and $p(▨▨▨) = 0$.

So, for the tree $T$ to be pruned, we could have a top-down tree pruning function TDTP described as the pseudo code, and a bottom-up tree pruning function BUTP.

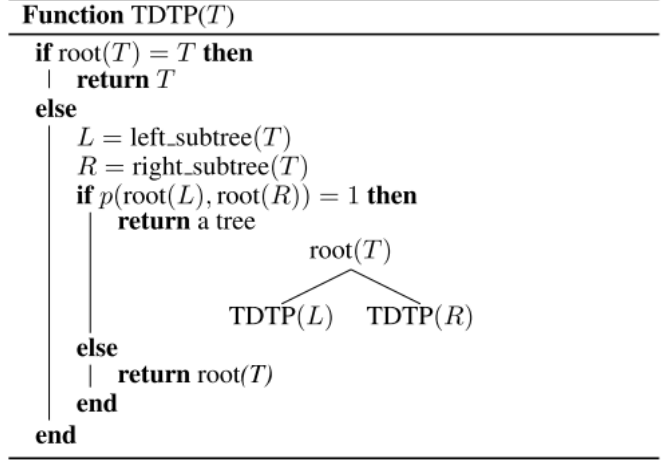

Fig. 5

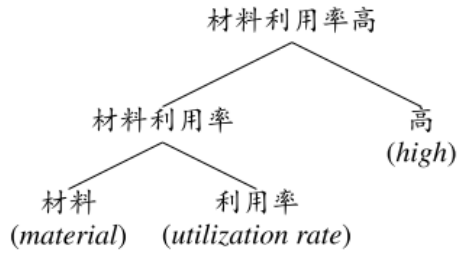

Fig. 6　A pruned tree for the binary tree in Fig. \ref{fig:tree}. The output for this pruned tree is "材料 利用率 高"

### 2.3 A Granularity Based Explanation

For the previous work like the CRF-based methods without the binary tree based framework, the outputted words can be directly determined by the function $b(c_i)$ and a threshold $t$:

The $t$-words of an input sentence $\overline{c_1 \ldots c_n}$ are any substrings like $\overline{c_i \ldots c_j}$, such that

$$\min\left(b(c_{i-1}), b(c_j)\right) > t > \max\left(b(c_i), \cdots, b(c_{j-1})\right)$$

**Definition**: The inequalities in this definition mean that there is a word boundary between $c_i$ and $c_{i+1}$ if and only if $b(c_i) > t$.

Notice that different results can be got with different thresholds. The greater the threshold is, the more coarse-grained the segmentation result is, which means there are lesser number of words in the output. The threshold $t$ can be seen as a parameter to control the output granularity.

It is better to store all the results with different granularity (by different thresholds). We can use an altered definition as:

**Definition**: The word candidates of an input sentence $\overline{c_1 \ldots c_n}$ are any substrings like $\overline{c_i \ldots c_j}$ such that $\min\left(b(c_{i-1}), b(c_j)\right) > \max\left(b(c_i), \cdots, b(c_{j-1})\right)$

The only difference is that there is no threshold $t$ in the inequalities. This definition is also natural. The explanation of this definition is that if the left and right boundaries of a string are more likely to be word boundaries than any character boundaries inside the string, this string may be a word.

The relation between the binary tree and the word candidates of an input sentence can be described as a proposition:

**Proposition**: A string is a word candidate if and only if it is in the binary tree of the corresponding sentence.

This proposition shows that the binary tree is a suitable representation to store all the word candidates with different granularity, which provides rich information for the tree pruning process to focus on the granularity problem.

## 3 Binary Tree Based Error Analysis

The statistics-based CWS algorithms are lack of error analysis methods. In the SIGHAN bake-off, the errors are only been divided into IV word errors and OOV word errors. Here we propose a new way to classify the errors for methods such as the CRF-based ones, and a way to investigate the performance without the granularity mismatch problem.

The granularity mismatch is an important cause for the errors. Since using binary trees is a way to maintain all the results for different granularity, the use of the binary trees can be helpful for the error analysis.

If an error is only caused by the granularity mismatch, the corresponding word in the gold standard should be found in the binary tree (It also should be the word candidate as we discussed in Section 2), although it is not in the output of the pruned tree. Otherwise, the corresponding word cannot be found in the binary tree.

According to this idea, in our framework, we divide the errors into tree errors and pruning errors, and the pruning errors can be caused by either over-pruning or less-pruning. We describe this as follows:

- **Tree error**. If a word in the gold standard cannot be found in the binary tree, it is called a tree error. It also means that the word is not in the output according to the threshold-based pruning method with any thresholds. This kind of errors cannot be corrected in the tree pruning step in our framework.
- **Pruning error**. If a word in the gold standard and can be found in the binary tree but it is not in the output, it is called a pruning errors. This kind of errors is caused by the granularity mismatch and could be corrected in the tree pruning step. These errors can be further divided into two subcategories:
    - **Over-pruning error**. If a word in the gold standard is pruned by the tree pruning function, it is called a over-pruning error. This is because of that the segmentation granularity for this word is too coarse.
    - **Less-pruning error**. If a word in the gold standard and its children are not pruned by the tree pruning function, it is called a less-pruning error. This is because of that the segmentation granularity for this word is too fine.

Both IV and OOV words may have tree errors and pruning errors. This is a new dimension to describe the errors besides the IV-OOV-based classification.

In order to estimate the upper bound of the performance for the tree pruning step, we define the **oracle pruning function** based on the gold standard:

$$p_{\text{oracle}}(\overline{c_i \ldots c_m}, \overline{c_{m+1} \ldots c_j})$$
$$= \begin{cases} 1, & \text{there is no word boundary between } c_m \text{ and } c_{m+1} \text{ in the gold standard;} \\ 0, & \text{otherwise.} \end{cases}$$

([x])

This function can be seen as a perfect pruning. By this pruning function, we can investigate the performance without the granularity mismatch problem for both ``mono corpus'' CWS and cross-corpus CWS.

## 4 An SVM-based Tree Pruning Function

Here we introduce a more sophisticated SVM-based function $p_{\text{SVM}}$ for the tree pruning in our framework. A state-of-the-art CRF-based model is used as the tree building function.

We need two training sets. The training set $\mathbf{S}_b$ is used to train a CRF-based CWS model. The training set $\mathbf{S}_p$ is used to train the SVM-based tree pruning model.

We need to train an SVM model as the binary pruning function $p_{\text{SVM}}()$. Sentences in $\mathbf{S}_p$ will first be parsed to binary trees by the trained CRF-based tree building function. Any input pairs for $p(\overline{c_i \ldots c_m}, \overline{c_{m+1} \ldots c_j})$ that $0.95 > b(c_m) > 0.05$ are with less confidence for the CRF model and are used as the samples to train the SVM model. The oracle pruning $p_{\text{oracle}}()$ values are used as the answers.

There are some notations for describing the features. For the pruning function $p_{\text{SVM}}(\overline{c_i \ldots c_m}, \overline{c_{m+1} \ldots c_j})$, we have $\boldsymbol{l} = \overline{c_i \ldots c_m}\$$, $\boldsymbol{r} = \overline{c_{m+1} \ldots c_j}$ and $\boldsymbol{m} = \overline{c_i \ldots c_j}$. We also have $\boldsymbol{l_{-1}} = \overline{c_u \ldots c_{i-1}}$ that $b(c_{u-1}) > 0.5$ and $b(c_k) < 0.5$ for $k = u, \ldots, i-2$, which is the word on the left side of $\boldsymbol{l}$ according to the CRF-based tree building function. Similarly we have the string $\boldsymbol{r_{+1}}$. The operator $\|\boldsymbol{s}\|$ returns the frequency of the string $\boldsymbol{s}$ in a corpus. The operator $\|\boldsymbol{s}\|_{\text{tree}}$ returns the frequency of the string $\boldsymbol{s}$ in the binary trees.

The features for the SVM are described as follows:

**CRF-based features**: The CRF-based features include the probability $P(t_i = \text{S} \vee t_i = \text{E}|\boldsymbol{s})$, and the marginal probabilities, $P(t_k = t|\boldsymbol{s})$ for $k \in \{m, m+1\}$ and $t \in \{\text{B}, \text{M}, \text{E}, \text{S}\}$.

**Length-based features**: Each character is one syllable in Chinese. Since syllable-length is a factor for Chinese word forming, we use binary features to represent the length-based information for the strings $\boldsymbol{l}$ and $\boldsymbol{r}$.

**Dictionary-based features**: We use the word unigrams and word bigrams in $\mathbf{S}_b$ to construct a unigram dictionary and a bigram dictionary. We use binary features to represent whether the strings $\boldsymbol{l}$, $\boldsymbol{r}$ and $\boldsymbol{m}$ are IV or OOV, respectively. Similar features are for the string pairs $(\boldsymbol{l_{-1}}, \boldsymbol{l})$, $(\boldsymbol{l}, \boldsymbol{r})$, $(\boldsymbol{r}, \boldsymbol{r_{+1}})$, $(\boldsymbol{l_{-1}}, \boldsymbol{m})$ and $(\boldsymbol{m}, \boldsymbol{r_{+1}})$.

**Association-measure-based features**: These features are designed to capture the global information of the strings. We define a chi$^2$-like function to measure the association between two strings. For the string pair $(\boldsymbol{l}, \boldsymbol{r})$, we set $a = \|\boldsymbol{m}\|$, $a + b = \|\boldsymbol{l}\|$, $a + c = \|\boldsymbol{r}\|$ and $a + b + c + d$ to be the total number of the characters. The frequencies are counted in training sets and the test set. The feature value is $(ad - bc)^2/(a+b)(a+c)(b+d)(c+d)$. Similar feature values can be calculated for the string pairs $(\boldsymbol{l_{-1}}, \boldsymbol{l})$, $(\boldsymbol{r}, \boldsymbol{r_{+1}})$, $(\boldsymbol{l_{-1}}, \boldsymbol{m})$ and $(\boldsymbol{m}, \boldsymbol{r_{+1}})$.

**Tree-based features**: These features are designed to capture the local information of the strings in the context. The idea is that if a string appears frequently in a document, it is likely to be a word. The frequencies are counted in either $\mathbf{S}_p$ or the test set. The used feature values are $\log\|\boldsymbol{m}\|_{\text{tree}}$ and

$\log\|\boldsymbol{m}\|_{\text{tree}} - \max_{\boldsymbol{m}'\in\text{pa}(\boldsymbol{m})} \log\|\boldsymbol{m}'\|_{\text{tree}}$, where $\text{pa}(\boldsymbol{m})$ is the set of all the parent nodes of $\boldsymbol{m}$ in the trees.

We use the CRF-based tree building function $b_{\text{CRF}}()$ and the SVM-based tree pruning function $p_{\text{SVM}}()$ for the test.

## 5 Experiments

### 5.1 Datasets and the Baseline

We use these four corpora of the SIGHAN bake-off 2005~\cite{emerson_second_2005} for our evaluation. They are free and widely used for the evaluation by most of the previous work.

The used measurements for the evaluation are the precision that $p = (\text{\# of words segmented correctly})/(\text{\# of words in the output})$, the recall that $r = (\text{\# of words segmented correctly})/(\text{\# of words in the gold standard})$ and the F_measure $= 2pr/(p+r)$. Besides, the OOV rate is also calculated for the analysis.

We use a state-of-the-art CRF-based method as the baseline, and to define the tree building function $b(c_i)$. The error analysis is also based on it. The CRF-based model is trained using the toolkit Pocket CRF[3]. The used feature templates for character $c_i$ are:

Table 1

| Types | Templates |
|---|---|
| unigram | $c_{i-1}t_i$, $c_i t_i$, $c_{i+1}t_i$ |
| bigram | $c_{i-2}c_{i-1}t_i$, $c_{i-1}c_i t_i$, $c_i c_{i+1}t_i$, $c_{i+1}c_{i+2}t_i$ |
| transfer | $t_{i-1}t_i$ |

### 5.2 Binary Tree Based Error Analysis

First, we investigate the numbers of different errors for the baseline method. The errors are not only divided into IV errors and OOV errors, but are also divided based on the binary trees as we discussed in the previous section.

Table 2 **Numbers of errors in different kinds for the CRF-based baseline for the SIGHAN bake-off 2005 corpora**

| | | AS | | CityU | | MSR | | PKU | |
|---|---|---|---|---|---|---|---|---|---|
| | | IV | OOV | IV | OOV | IV | OOV | IV | OOV |
| tree error | | 310 | 412 | 234 | 271 | 183 | 172 | 313 | 443 |
| pruning error | over-pruning | 2661 | 190 | 1166 | 420 | 2839 | 139 | 4287 | 227 |
| | less-pruning | 841 | 1039 | 92 | 263 | 268 | 593 | 309 | 590 |

The results are in Table \ref{tab:error_types}. Among the four different copora, the phenomena are similar. Tree errors are much less than the pruning errors. This indicates that the granularity mismatch is the most primary cause of the errors.

We can also see that there are more IV words for the over-pruning errors while there are OOV words for the less-pruning errors. This is due to the phenomenon that most of the OOV words are consist of IV words but not vice versa.

---
[3] http://pocket-crf-1.sourceforge.net/

Then, with the help of the oracle pruning function, we can see what performance we can get without the granularity mismatch problem.

Table 3  The OOV rate and the F measures of the threshold based pruning and the oracle pruning for four corpora

|  | AS | CityU | MSR | PKU |
|---|---|---|---|---|
| OOV rate | 4.3 | 7.4 | 2.6 | 5.8 |
| F measures for CRF | | | | |
| $p_{\text{threshold}_{0.5}}$ | 95.0 | 94.3 | 96.2 | 94.6 |
| $p_{\text{oracle}}$+TDTP | 99.1 | 98.0 | 99.5 | 98.9 |
| $p_{\text{oracle}}$+BUTP | 99.1 | 98.1 | 99.5 | 98.9 |

The F measures by the threshold based pruning function (the output will be the original output by the CRF-based baseline) and the oracle pruning function can be found in Table \ref{tab:baseline}. These results show that the upper bound of the F_measure for the best pruning function is quite high. We see that a better pruning function is useful to improve the performance.

Then we investigate the performances of cross-corpus CWS. The MSR and PKU corpus are both in simplified Chinese. We trained a CRF-based model using the training set of the MSR corpus and test it on the test set of the PKU corpus. This experiment is called `MSR to PKU'. Similar experiment `PKU to MSR' is also performed.

Table 4  The analysis of the performances for cross-corpus CWS

|  | MSR to PKU | | | PKU to MSR | | |
|---|---|---|---|---|---|---|
|  | precision | recall | F1 | precision | recall | F1 |
| CRF | 87.7 | 82.5 | 85.0 | 84.1 | 87.5 | 85.8 |
| +$p_{\text{oracle}}$+TDTP | 98.0 | 95.0 | 96.5 | 97.2 | 91.0 | 94.0 |
| +$p_{\text{oracle}}$+TDBU | 95.7 | 97.9 | 96.8 | 94.1 | 97.1 | 95.5 |

The results for these two cross-corpus CWS experiments are in Table \ref{tab:cross_corpus}. Though the F measures of the threshold based pruning function (i.e. the original CRF-based model) are much poor, the F measures of the oracle pruning function are still high. Especially for the `MSR to PKU', there are more than 98\% of the words in the gold standard which can be found in the binary trees. The morphological and syntax structures are the same in any corpora. The drop of the performance for the cross-corpus experiments are caused by the worse granularity mismatch problem.

### 5.3 SVM-based Tree Pruning

In order to compare to the previous works based on the same training sets in SIGHAN bake-off 2005 and avoid using any other resources, we divided the original training set into two parts. Nine tenths of them are used as the training set $\mathcal{S}_\textup{b}$ to train the CRF model for the tree building, while the rest one tenth are used as the training set $\mathcal{S}_\textup{p}$ to train the SVM model for the tree pruning.

We use LibSVM[4] for the training and testing for the SVM-based model, and use all the default

---
[4]  http://www.csie.ntu.edu.tw/~cjlin/libsvm/

parameters. Features are described in Section \ref{section:SVM}.

The results are in Table \ref{tab:svm}. The first and second rows show the F measures of the original CRF-based baseline with 100% and 90% of the training set, respectively. The third row shows the F measures of the method that uses the SVM-based pruning function.

We see that the performance only drops a little when we reduce 10\% of the training data for the CRF model. After using them for the training of the more sophisticated SVM-based pruning model, the performances increase. If we define the error rate as $1-\textup{F-measure}$, the error reduction is about 10\% and up to 20\%, which is significant for the CWS evaluation. Other experiments also show that all kinds of features help the performance.

Table 5   **A comparison between the baseline CRF-based method and the method using SVM-based pruning on the F measures**

|  | AS | CityU | MSR | PKU |
|---|---|---|---|---|
| 100%CRF | 95.0% | 94.3% | 96.2% | 94.6% |
| 90%CRF | 95.0% | 94.1% | 96.1% | 94.4% |
| $+p_{\text{SVM}}$ | 95.4% | 94.7% | 97.1% | 95.0% |

## 6 Conclusions

We proposed a binary tree representation for the structures of the unclear part between the Chinese morphology and syntax. We also proposed a simple binary tree based two-step framework for CWS, namely tree building and tree pruning. Previous models for CWS can be employed in this framework.

The binary tree representation provides a quantitative error analysis method for CWS, by which we see that the granularity mismatch problem is the primary cause of the errors for CWS and cross-corpus CWS.

We also illustrated with an SVM-based tree pruning model for the Step 2, and reduce the error rate up to 20\% from a state-of-the-art CRF-based baseline.

The definition of Chinese word is not clear even for the linguists\cite{xue_defining_2001}. The disagreements of the segmentation standard between different corpora such as the disagreement between MSR and PKU corpus are mainly on the granularity. Moreover, applications such as machine translation and information retrieve need CWS models with different granularity.

Our binary tree representation not only provides a way to improve the performance of CWS, but also provides a way to solve these problems. We can have a unified tree building function and different tree pruning functions for different corpora and applications with different granularity.